\documentclass[lettersize,journal]{IEEEtran}
\usepackage{amsmath,amsfonts,amssymb}
\usepackage{multirow}
\usepackage{siunitx}
\usepackage{algorithmic}
\usepackage{algorithm}
\usepackage{array}
\usepackage[caption=false,font=normalsize,labelfont=sf,textfont=sf]{subfig}
\usepackage{textcomp}
\usepackage{stfloats}
\usepackage{hyperref}
\usepackage{url}
\usepackage{verbatim}
\usepackage{graphicx}
\graphicspath{ {./images/} }
\usepackage{cite}
\usepackage{subeqnarray}
\usepackage{cases}
\hypersetup{hypertex=true,
            colorlinks=true,
            urlcolor=black,
            linkcolor=blue,
            anchorcolor=blue,
            citecolor=blue}
\newcommand\blfootnote[1]{%
\begingroup 
\renewcommand\thefootnote{}\footnote{#1}%
\addtocounter{footnote}{-1}%
\endgroup 
}

\begin{document}

\title{Motion Planning and Control of A Morphing Quadrotor in Restricted Scenarios}

\author{Guiyang Cui$^{1}$, Ruihao Xia$^{1}$, Xin Jin$^{2}$, and Yang Tang$^{*,1}$, \textit{Fellow}, \textit{IEEE}

\thanks{This paper was produced by the IEEE Publication Technology Group. They are in Piscataway, NJ.}
\thanks{Manuscript received April 19, 2021; revised August 16, 2021.}}


\twocolumn[{%
\renewcommand\twocolumn[1][]{#1}%
\maketitle
 \vspace*{-1.5cm}
\begin{figure}[H]
\hsize=\textwidth
    \includegraphics[width=1\textwidth]{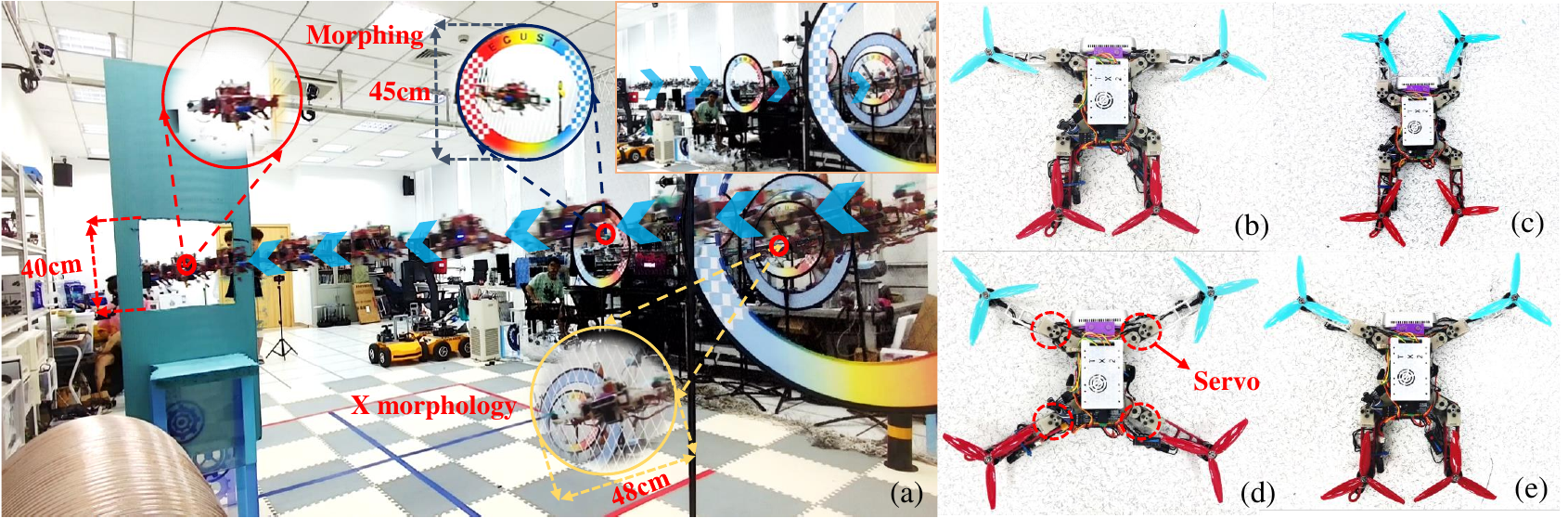}\\
    \vspace{-0.6cm}
    \caption{(a) The morphing quadrotor is maneuvering through complex and restricted environments. (b) Close-range vertical surface inspection in a T-shaped configuration. (c) Traversal of narrow gaps or object grasping and transportation in an H-shaped configuration. (d) Normal flight in an X-shaped configuration. (e) Minor hovering thrust in a Y-shaped configuration. A high-resolution video is available at: \url{https://youtu.be/0_GC4fv9Eok}.} 
    \label{fig1} 
\end{figure}
\setcounter {figure}{1}
}]

\blfootnote{\hspace{-0.115cm} $^{1}$The authors are with the Key Laboratory of Smart Manufacturing in Energy Chemical Process Ministry of Education, East China University of Science and Technology, Shanghai, 200237, China.}
\blfootnote{\hspace{-0.0cm}$^{2}$The author is with the Research Institute of Intelligent Complex Systems, Fudan University, Shanghai, 200433, China.}
\blfootnote{\hspace{-0.0cm}*Corresponding author’s e-mail: yangtang@ecust.edu.cn (Y.Tang)
}

\vspace*{-0.3cm}
\begin{abstract}
Morphing quadrotors with four external actuators can adapt to different restricted scenarios by changing their geometric structure. However, previous works mainly focus on the improvements in structures and controllers, and existing planning algorithms don't consider the morphological modifications, which leads to safety and dynamic feasibility issues. In this paper, we propose a unified planning and control framework for morphing quadrotors to deform autonomously and efficiently. The framework consists of a \textit{milliseconds}-level spatial-temporal trajectory optimizer that takes into account the morphological modifications of quadrotors. The optimizer can generate full-body safety trajectories including position and attitude. Additionally, it incorporates a nonlinear attitude controller that accounts for aerodynamic drag and dynamically adjusts dynamic parameters such as the inertia tensor and Center of Gravity. The controller can also online compute the thrust coefficient during morphing. Benchmark experiments compared with existing methods validate the robustness of the proposed controller. Extensive simulations and real-world experiments are performed to demonstrate the effectiveness of the proposed framework.
\end{abstract}


\section{Introduction}
\IEEEPARstart{I}{n} recent years, quadrotors have been widely applied for exploration and rescue expeditions \cite{ref1}. However, their fixed geometric structure limits the ability to navigate through narrow spaces smaller than their size, including pipes, mines, caves, and other restricted scenarios.
Traditional quadrotors require aggressive flight maneuvers with large attitude angles to pass through these narrow spaces \cite{ref2}, which makes it difficult to maintain accurate control at high velocities and angular accelerations.
In contrast, morphing quadrotors can adapt to specific restricted scenarios by employing four external actuation mechanisms to fold arms and reduce their size \cite{ref3,ref4,ref5,ref6}, utilizing passive hinges for volume compression\cite{ref7}, decreasing body width through rotating arms\cite{ref8}, or achieving shape reduction with telescopic arms\cite{ref9}.
Morphing quadrotors in \cite{ref3} offers advantages over \cite{ref7,ref8,ref9}  by exhibiting lower control difficulty and adaptability to different task scenarios, including traversing narrow gaps \cite{ref7,ref8}, grasping and transporting objects \cite{ref10,ref11}. Therefore, we adopt the structure proposed in \cite{ref3}, as shown in Fig. \ref{fig1}(b-e).
However, this work is limited to the structure and controller design. Compared to conventional quadrotors, motion planning and control of morphing quadrotors face the following challenges:

\begin{enumerate}
\item{When navigating through restricted environments, it becomes challenging to account for morphological modifications that generate collision-free trajectories for the full body, including position and attitude in real-time.}
\item{Sudden changes in the inertia tensor and Center of Gravity (CoG), along with the uncertain control allocation during morphing, result in compromised control accuracy.}

\end{enumerate}

In this work, we focus on the motion planning and control of morphing quadrotors in restricted scenarios, which has been rarely explored in previous works. With the proposed framework, the morphing quadrotor is capable of autonomously deforming and traversing restricted obstacles in complex environments safely and swiftly (as illustrated in Fig. \ref{fig1}(a)). The obstacles are composed of two circles with inner diameters of \SI{45}{\centi\meter} and \SI{80}{\centi\meter}, as well as a square notch with a side length of \SI{40}{\centi\meter}. The overall software architecture is shown in Fig. \ref{fig2}.
We also conduct extensive experiments in challenging real-world environments to demonstrate the performance and robustness of the proposed framework. Finally, experiments compared with existing  methods\cite{ref3,ref4} for continuous variable-shape flight effectively validate the superior control accuracy of our work. 

Our contributions can be summarized as follows:
\begin{figure*}[!t]
    \hspace{1.3cm}\includegraphics[width=0.85\textwidth]{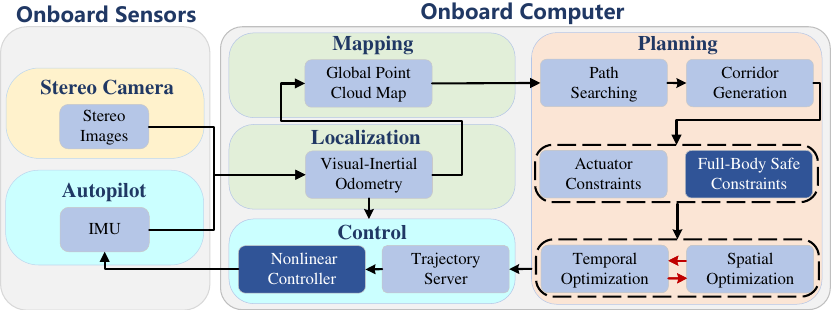}
    \vspace{-0.2cm}
    \caption{Software architecture. Perception, planning, and control modules are running onboard. Innovative planning and control strategies designed for morphing quadrotors are represented by the deep blue areas. More details are included in the contributions.}
    \vspace{-0.4cm}
    \label{fig2}
\end{figure*}  
\begin{enumerate}
\item{A \textit{milliseconds}-level spatial-temporal trajectory optimizer is proposed. Compared with existing planning algorithms for traditional quadrotors\cite{ref12}, it takes into account the morphological modifications of quadrotors. Full-body collision-free trajectories including position and attitude can be generated in real-time.} 
\item{A unified nonlinear attitude controller for morphing quadrotors is proposed. Compared with existing controllers for morphing quadrotors \cite{ref3,ref4}, we consider the nonlinear aerodynamic drag to avoid significant distortion of the linearized model. The proposed controller can adjust the time-varying parameters, including the inertial tensor, CoG, and thrust coefficient during morphing to achieve a more stable flight.}
\item{Various controller comparisons show the advantages of the proposed method in control accuracy and stability. Extensive simulations and real-world experiments in different restricted scenarios effectively verify the effectiveness of the proposed framework.}
\end{enumerate}

\section{RELATED WORK}
Several works\cite{ref3,ref4,ref5,ref6,ref10,ref13,ref14} present control systems for morphing quadrotors with four external actuators. In real-world experiments, Derrouaou \textit{et al}. \cite{ref4}, and Kim \textit{et al}. \cite{ref10} conducted flight or grasping tests using linear PID controller. Falanga \textit{et al}. \cite{ref3} performed experiments on hovering, grasping, and exploration tasks using Linear Quadratic Regulator (LQR) controller. Fabris \textit{et al}. \cite{ref13} further investigated the effects of overlapping between the propellers and the vehicle body. Furthermore, Hu \textit{et al}. \cite{ref14} achieved stable hovering in the process of deformation control for morphing drones through Reinforcement Learning. However, these works don't consider the effect of nonlinear aerodynamic drag in the controller design process, resulting in lower control accuracy. In addition, the lack of motion planning integration limits the ability of quadrotors to achieve autonomous and safe flight in restricted environments.

In simulation experiments, Derrouaou \textit{et al}. \cite{ref4} utilized Particle Swarm Optimization to design an adaptive controller for trajectory tracking experiments. Papadimitriou \textit{et al}. \cite{ref5} employed Nonlinear Model Predictive Control to enable the drone to pass through narrow mine entrances. Butt \textit{et al}. \cite{ref6} took into account the attitude control and geometric tracking of a morphing quadrotor using the particular orthogonal group in three dimensions SO(3), and conducted simulation experiments involving multiple deformations and crossings. However, these works still lack real-world tracking experiments to validate the effectiveness of their algorithms.

In the proposed framework, we develop a nonlinear attitude controller based on differential flatness, which dynamically computes the inertia tensor, CoG, and control allocation during morphing. We utilize the weighted A* algorithm to generate a collision-free path and then construct the Safe Flight Corridors (SFC) comprising a series of convex polyhedra. Furthermore, the polyhedral constraints surrounding the full body are dynamically constructed during the deformation process. Finally, we employ MINCO \cite{ref12} to eliminate constraints exactly without introducing extra local minima.

\section{Dynamic Model}
\begin{figure}[!t]
    \hspace{0.7cm}\includegraphics[width=0.4\textwidth]{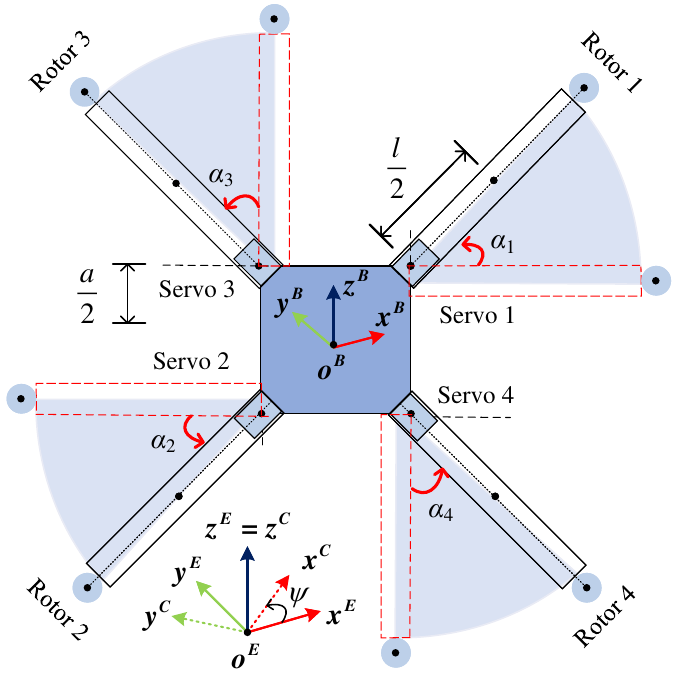}
    \caption{Deformation scheme of our morphing quadrotor.} 
    \label{fig3}
    \vspace{-0.3cm}
\end{figure}

In this work, we make use of bold italic letters to denote vectors or matrices (e.g., \(\boldsymbol{v, M}\)), otherwise they refer to scalars.
As shown in Fig. \ref{fig3}, let $\left\{\boldsymbol{x}^{E},\boldsymbol{y}^{E},\boldsymbol{z}^{E}\right\}$ and $\left\{\boldsymbol{x}^{B},\boldsymbol{y}^{B},\boldsymbol{z}^{B}\right\}$ represent the orthogonal bases in the world coordinate system \(\mathbf{E}\) and the body coordinate system \(\mathbf{B}\), respectively. The body frame is defined at the geometric center of the quadrotor. We denote \(\boldsymbol{\alpha_{i}} ({i}=1,2,3,4)\) as the angles of the arms rotating around the \(z\) axis. During the deformation process, let \(\boldsymbol{p}_{E}, \boldsymbol{p}_{C}\) and \(\boldsymbol{p}\) be the center of gravity positions of the initial, offset, and deformation in the world frame. The velocity \(\boldsymbol{v}\) and orientation \(\boldsymbol{q}\) remain the same as in the previous condition. Additionally, let \(\boldsymbol{R},\boldsymbol{\omega}\) represent the rotation matrix and the angular velocity in the body frame. 

Inspired by \cite{ref15}, the system dynamic equations for the morphing quadrotor considering aerodynamic drag can be obtained as follows:
\begin{subequations}
\begin{numcases}{}
\boldsymbol{p} =\boldsymbol{p}_{E}-\boldsymbol{p}_{C}, \label{eq1A} \\
\dot{\boldsymbol{p}} =\boldsymbol{v},  \label{eq1B} \\
\dot{\boldsymbol{v}} =-g \boldsymbol{z}^{E}+\boldsymbol{R} \frac{f}{m_{t}} \boldsymbol{z}^{E}-\boldsymbol{R} \boldsymbol{D} \boldsymbol{R}^\mathrm{T} \boldsymbol{v}, \label{eq1C}\\
\dot{\boldsymbol{R}} =\boldsymbol{R} \hat{\boldsymbol{\omega}}, \label{eq1D} \\
\dot{\boldsymbol{\omega}} =\boldsymbol{J}^{-1}\left(\boldsymbol{\tau}-\boldsymbol{\omega} \times \boldsymbol{J} \boldsymbol{\omega} - \boldsymbol{A} \boldsymbol{R}^\mathrm{T} \boldsymbol{v}-\boldsymbol{B} \boldsymbol{\omega}\right), \label{eq1E}
\end{numcases}
\end{subequations}
where \(m_{t}\) and \(f\) are the total mass and collective thrust, \(\boldsymbol{z}^{E} = \left [0,0,1\right]^\mathrm{T}\) is the unit vector in the inertial frame, \(\boldsymbol{D} = \mathrm{diag}\left ({d}_{x},d_{y},d_{z}\right)^\mathrm{T}\) is a constant diagonal matrix normalized by mass that is associated with rotor drag coefficient, \(\boldsymbol{\tau} = \left [\tau_{x},\tau_{y},\tau_{z}\right]^\mathrm{T}\in \mathbb{R}^{3}\) represents the moment generated by the propeller on the body axis, while A and B are constant
matrices corresponding to self-rotation damping moment and
motion translational moment, respectively. The inertia matrix \(\boldsymbol{J}\) and the gravity offset \(\boldsymbol{p}_{C}\) can be obtained using the calculation method described in \cite{ref3}.

Let \(c_{M}\),\(c_{T}\) be the torque and thrust coefficient of \(i\)-th motor. \(\boldsymbol{\Omega}_{i}\) and \(\boldsymbol{l}_{i}=\left[l_{xi},l_{yi},l_{zi}\right]^\mathrm{T}\) are the rotational speed of the \(i\)-th motor and its position in the body frame. The collective thrust \(f\) and torque \(\boldsymbol{\tau}\) are expressed by
\begin{equation}
\label{eq2}
\left[\begin{array}{l}
f \\
\boldsymbol{\tau}
\end{array}\right]=\boldsymbol{M}_{C} \boldsymbol{U},
\end{equation}
where \(\boldsymbol{U}=\left[c_{t}{\Omega}_{1}^{2},c_{T}{\Omega}_{2}^{2},c_{T}{\Omega}_{3}^{2},c_{T}{\Omega}_{4}^{2}\right]^\mathrm{T}\) represents the thrust generated by each rotor, \(\boldsymbol{M}_{C}\) is the control allocation matrix during morphing. Assuming that the center of gravity offset is \(\boldsymbol{p}_{C}=\left[x_{C},y_{C},z_{C}\right]^\mathrm{T}\) at this time, then \(\boldsymbol{M}_{C}\) is as follows:
\begin{equation}
\label{eq3}
\boldsymbol{M}_{C}=\left[\begin{array}{cccc}
1 & \left(y_{C}-l_{y 1}\right) & \left(l_{x 1}-x_{C}\right) & c_{M}/c_{T}  \\
1 & \left(y_{C}-l_{y 2}\right) & \left(l_{x 2}-x_{C}\right) & c_{M}/c_{T}  \\
1 & \left(y_{C}-l_{y 3}\right) & \left(l_{x 3}-x_{C}\right) & -c_{M}/c_{T} \\
1 & \left(y_{C}-l_{y 4}\right) & \left(l_{x 4}-x_{C}\right) & -c_{M}/c_{T} 
\end{array}\right]^{\mathrm{T}}.
\end{equation}

\section{Motion Planning For Morphing Quadrotors}

In this section, we introduce a motion planning optimizer for a morphing quadrotor that takes into account the morphological modifications of quadrotors. The optimizer consists of two main components: the front-end, which comprises the construction of Safe Flight Corridors (SFC) using weighted A* search, and the back-end, which utilizes the MINCO trajectory optimization method with the full-body safety constraints.

\subsection{Full-Body Safety Constraints}
\begin{figure}[!t]
    \hspace{-0.2cm}\includegraphics[width=0.5\textwidth]{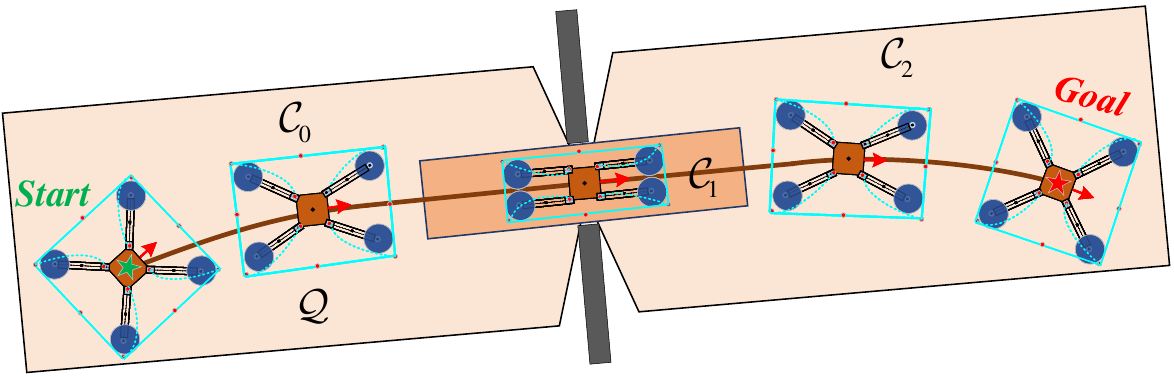}
    \vspace{-0.6cm}
    \caption{The morphing quadrotor traverses a narrow gap in a planar diagram. Polyhedral \(\mathcal{Q}\) is determined by the quadrotor's actual shape, as illustrated in the cyan box. \(\mathcal{C}_{k}\) denotes convex polyhedral units, depicted by the orange region. }
    \label{fig4}
    \vspace{-0.3cm}
\end{figure}

Firstly, We perform path searching on the constructed 3D point cloud map using the weighted A* algorithm \cite{ref16}. Compared to the A* algorithm, the weighted A* algorithm exhibits the characteristic of greedy search, enabling faster identification of the optimal path 
\(P=\left\{\boldsymbol{p}^{*}_{0},\boldsymbol{p}^{*}_{1},\ldots,\boldsymbol{p}^{*}_{n}\right\}\). 
Following the identification of path points, we construct a series of convex polyhedral-shaped SFC \(\mathcal{F}\left(P\right)=\left\{\mathcal{C}_{k}|k=1,2,\ldots, K \right\}\), as depicted by the orange region in Fig. \ref{fig4}. These convex units are stipulated to be interconnected in a local sequential manner, defined as:
\begin{equation}
    \label{4}
    \begin{aligned}
    \mathcal{C}_{k}&=\left\{\boldsymbol{p} \in \mathbb{R}^{3} \mid\left(\boldsymbol{p}-\boldsymbol{r}_{k}^{j}\right)^\mathrm{T} {\boldsymbol{n}}_{k}^{j} \leq 0 \right\}, \\
    \forall j &\in \left \{1,\ldots,M\right\},\forall k\in \left \{ 1,\ldots,K \right \}.
    \end{aligned}
\end{equation}
Each convex polyhedral unit \(\mathcal{C}_{k}\) consists of \(M\) hyperplanes, with each hyperplane being defined by an inward-facing normal vector \({\boldsymbol{n}}_{k}^{j}\) and a point \(\boldsymbol{r}_{k}^{j}\) on the plane.

To achieve the full body into collision-free space \(\mathcal{F}\) instead of just focusing on the CoG of the morphing quadrotor. We utilize polyhedral \(\mathcal{Q}(t)\) to represent the body of the quadrotor.
\begin{equation}
    \label{eq5}
    \begin{aligned}
    \mathcal{Q}_{k}(t)=\left\{\boldsymbol{q} \in \mathbb{R}^{3} \mid \boldsymbol{q}=\boldsymbol{R} \hat{\boldsymbol{q}}_{k}(t)+\boldsymbol{p}_{k}(t) \right\},  \\
    \left({\boldsymbol{n}}_{k}^{j}\right)^\mathrm{T}\left(\boldsymbol{q}-\boldsymbol{r}_{k}^{j}\right) \leq 0 , \forall k\in \left \{ 1,\ldots,K \right \}, 
    \end{aligned}
\end{equation}
where \(\boldsymbol{q} \in \mathcal{Q}_{k}(t)\) represents the position of any point on the overall drone in the world coordinate system,  \(\boldsymbol{p} \in \mathbb{R}^{3}\) and \(\boldsymbol{R} \in \mathrm{SO}{\left(3\right)}\) represent the position of the center of gravity and the rotation matrix, respectively. \(\hat{\boldsymbol{q}}=\left[\hat{q}_{x},\hat{q}_{y},\hat{q}_{z}\right]^\mathrm{T} \in \hat{\mathcal{Q}}\), and the value of \(\hat{\mathcal{Q}}\) depends on the actual shape of the drone, as shown in the cyan rectangular box in Fig. \ref{fig4}.
In this work, we choose to model the drone as a cube, with half of its dimensions represented by \(r\), \(w\), and \(h\) for length, width, and height, respectively. This approach only requires considering eight vertices.
\begin{equation}
    \label{eq6}
    \hat{\mathcal{Q}}=\left\{\hat{\boldsymbol{q}}_{v}=[ \pm r \pm w \pm h]^\mathrm{T}, v=1,2, \ldots, 8\right\}.
\end{equation}

However, during the deformation process, the length and width of the quadrotor will vary with the angle of the servomotors. Then, half of the length and width of the quadrotor can be expressed as follows:
\begin{equation}
    \label{eq7}
    \left\{\begin{aligned}
    r_{new}  &=\left(a+l \sin \alpha_{i}\right) / 2 \\
    w_{new} &=\left(a+l \cos \alpha_{i}\right) / 2
    \end{aligned} , \forall i \in\{1,2,3,4\}\right.,
\end{equation}
where \(a\) and \(l\) represent the length of arm and central body. Thus, dynamic polyhedral constraints are constructed in real-time to achieve full-body collision-free.

Currently, several existing works on trajectory planning based on SFC \cite{ref17,ref18,ref19}, and we utilize the RILS method from \cite{ref19} to generate the SFC in our work. Subsequently, we integrate the SFC, along with the constructed drone model, as user-defined safety constraints applied to the spatial-temporal deformation of the MINCO trajectory class \cite{ref12}.
\vspace{-0.1cm}
\subsection{Problem Formulation}
In this paper, we adopt \(\mathfrak{T}_{\mathrm{MINCO}}\) \cite{ref12}, a minimum control effort polynomial trajectory class defined as:
\vspace{-0.2cm}
{\setlength\abovedisplayskip{0.45cm}
\setlength\belowdisplayskip{0.3cm}
\begin{equation}
\label{eq8}
\begin{array}{c}
\mathfrak{T}_{\mathrm{MINCO}}=\left\{\boldsymbol{p}(t):[0, T] \mapsto \mathbb{R}^{m} \mid \mathbf{c}=\mathcal{M}(\mathbf{q}, \mathbf{T})\right., \\ \\ 
\left.\mathbf{q} \in \mathbb{R}^{m(K-1)}, \mathbf{T} \in \mathbb{R}_{>0}^{M}\right\},
\end{array}  
\end{equation}}
\hspace{-0.2cm}where an \(m\)-dimensional trajectory \(\boldsymbol{p}(t)\) is represented by a piece-wise polynomial of \(K\) pieces and \(N = 2s - 1\) degree. We define the intermediate waypoints \(\mathbf{q}=\left(q_{1},\ldots,q_{K-1}\right)\) and time allocation vector \(\mathbf{T}=\left(T_{1},\ldots,T_{K}\right)^\mathrm{T}\).

To ensure the smoothness of the position trajectory \(\boldsymbol{p}(t)\) and meet the requirements of dynamic feasibility, 
We measure the smoothness quality of the trajectory by the magnitude of \(\left\|\boldsymbol{p}^{(s)}(t)\right\|_{2}^{2}\) (\(s=3\) in our experiments).
The goal of our trajectory optimization is to find a full-state trajectory \(\boldsymbol{q}(t)\)
within the SFC that connects the start and end positions, as shown by the brown line in Fig. \ref{fig4}. The expression for optimizing the full-state trajectory of the morphing quadrotor is given as follows:
\vspace{-0.2cm}
{\setlength\abovedisplayskip{0.4cm}
\setlength\belowdisplayskip{0.2cm}
\begin{subequations}\label{eq9}
    \begin{align}
    \hspace{-0.2cm}
    \min _{\boldsymbol{p}(t),t_{K}} & \mathcal{J}_{o} = \int_{0}^{t_{K}}\left\|\boldsymbol{p}^{(s)}(t)\right\|_{2}^{2} d t+\rho_{T} t_{K},  \label{eq9A}\\
    \text { s.t. } & \boldsymbol{p}(t)=\mathbf{c}_{k}^{\mathrm{T}}\left(\mathbf{q,T}\right)\boldsymbol{\beta}\left(t-t_{k-1}\right), \forall t \in\left[t_{k-1}, t_{k}\right], \label{eq9B}\\
    & \boldsymbol{p}^{[0: s-1]}(0)=\boldsymbol{d}_{0}, \boldsymbol{p}^{[0: s-1]}\left(t_{K}\right)=\boldsymbol{d}_{g}, \label{eq9C}\\
    & t_{k-1}<t_{k}, \forall 1 \leq k \leq K, \label{eq9D}\\
    & \left\|\boldsymbol{p}^{(1)}(t)\right\|_{2}^{2} \leq v_{\max }^{2}, \label{eq9E}\\
    & \left\|\boldsymbol{\omega}^{(2)}(t)\right\|_{2}^{2} \leq \omega_{\max }^{2}, \label{eq9F}\\
    & \boldsymbol{q}(t)=\boldsymbol{q}\left(t-t_{k-1}\right) \in \mathcal{C}_{k}, \forall 1 \leq k \leq K, t \in\left[t_{k-1}, t_{k}\right], \label{eq9G} 
    \end{align}
\end{subequations}}
\hspace{-0.15cm}where equation \eqref{eq9A} represents the cost function that balances smoothness and maneuverability. Equation \eqref{eq9B} denotes the coefficient matrix of the polynomial trajectory \(\mathbf{c}_{k}\in \mathbb{R}^{2s \times 3}\) for each segment depending on \(\left(\mathbf{q,T}\right)\), and using \(\boldsymbol{\beta} \left(t\right)=\left[1,t,\ldots,t^{2s-1}\right]^\mathrm{T}\) as the time basis function. Equation \eqref{eq9C} represents the constraints on the initial and final states of the trajectory. Equation \eqref{eq9D} ensures that the time allocation for each trajectory segment is non-zero. Equations \eqref{eq9E} and \eqref{eq9F} represent the actuator constraints, primarily imposing safety constraints on the velocity \(\boldsymbol{p}^{(1)}(t)\) and body angular velocity \(\boldsymbol{\omega}\). Equation \eqref{eq9G} represents the safety constraints, ensuring that the full body remains within the SFC to avoid collisions.

The continuous-time constraints of (\ref{eq9E}--\ref{eq9G}) can be relaxed to an integral penalty term. As a consequence, the above optimization in \eqref{eq9} is converted into an unconstrained optimization problem.
\begin{equation}
    \label{eq10}
    \hspace{-0.5cm}
    \begin{array}{rl}\min _{\boldsymbol{p}_{k}{(t)}}& \mathcal{J}=\sum_{k=1}^{K}\left(\int_{0}^{T_{k}}\left\|\boldsymbol{p}_{k}^{(s)}(t)\right\|_{2}^{2} d t+\rho_{T} ({\left \| \mathbf{T}  \right \| })\right) \\ & +\underbrace{\rho_{\text {v }} \sum_{k=1}^{K} \int_{0}^{T_{k}} \mathcal{L}\left(\left\|\boldsymbol{p}_{k}^{(1)}(t)\right\|_{2}^{2}-v_{\max }^{2}\right) \mathrm{d} t}_{\text {velocity feasibility penalty }} \\& +\underbrace{\rho_{\text {w }} \sum_{k=1}^{K} \int_{0}^{T_{k}} \mathcal{L}\left(\left\|\boldsymbol{\omega}_{k}(t)\right\|_{2}^{2}-\omega_{\max }^{2}\right) \mathrm{d} t}_{\text {angular feasibility penalty }} \\& +\underbrace{\rho_{\mathrm{c}} \sum_{k=1}^{K} \int_{0}^{T_{k}} \sum_{j=1}^{M}\mathcal{L}((\boldsymbol{n}_{k}^{j})^\mathrm{T}(\boldsymbol{q}(t)-\boldsymbol{r}_{k}^{j}))  \mathrm{d} t}_{\text {collision-free penalty }},\end{array}
\end{equation}
where \(\mathcal{L}\left({x}\right)=\mathrm{max} \left(x,0\right)^{3}\) represents the cubic penalty term, \(\rho_{T}\) denotes the time regularization weight, and \(\rho_{v}\), \(\rho_{\omega}\), and \(\rho_{c}\) correspond to the penalty weights for maximum velocity, maximum angular rate, and collision-free. Finally, the unconstrained optimization problem can be solved using the Quasi-Newton method (LBFGS\footnote{\href{https://github.com/ZJU-FAST-Lab/LBFGS-Lite}{https://github.com/ZJU-FAST-Lab/LBFGS-Lite}}).
\vspace{-0.3cm}
\subsection{Gradient Calculation}
To solve the unconstrained optimization problem \eqref{eq10}, we need to obtain the gradients of all objective functions.

\subsubsection{Time-Regularized Smoothness \(\mathcal{J}_{o}\)}
The gradients of \(\partial \mathcal{J}_{o}/ {\partial {\mathbf{c}}}\) and \(\partial \mathcal{J}_{o}/ {\partial {\mathbf{T}}}\) are calculated as
\begin{subequations}\label{eq12}
\begin{align}
\frac{\partial \mathcal{J}_{o}}{\partial c_{k}} & = 2\left(\int_{0}^{T_{k}} \boldsymbol{\beta}^{(3)}(t) \boldsymbol{\beta}^{(3)}(t)^{\mathrm{T}} \mathrm{d} t\right) c_{k}, \label{17A}\\
\frac{\partial \mathcal{J}_{o}}{\partial T_{k}} & = c_{k}^{\mathrm{T}} \boldsymbol{\beta}^{(3)}\left(T_{k}\right) \boldsymbol{\beta}^{(3)}\left(T_{k}\right)^{\mathrm{T}} c_{k}+\rho_{T} .\label{17B}
\end{align}
\end{subequations}
\subsubsection{Continuous-Time Constraints \(\mathcal{J}_{v},\mathcal{J}_{\omega},\mathcal{J}_{c}\)}
The penalty function of dynamic feasibility terms (\(\mathcal{J}_{v},\mathcal{J}_{\omega}\)) and collision-free term \(\mathcal{J}_{c}\) can be represented as: \(\forall j \in\{1, \ldots,{M}\}\)
\begin{equation}
    \label{eq11}
    \left\{\begin{aligned}
    \mathcal{G}_{v}(t) &= \left\|\boldsymbol{p}_{k}^{(1)}(t)\right\|_{2}^{2}-v_{\max }^{2}, & \forall t \in\left[0, T_{k}\right], \\
    \mathcal{G}_{\omega}(t) &= \left\|\boldsymbol{\omega}_{k}(t)\right\|_{2}^{2}-{\omega}_{\max }^{2}, &\forall t \in\left[0, T_{k}\right], \\
    \mathcal{G}_{c}(t)&=(\boldsymbol{n}_{k}^{j})^\mathrm{T}(\boldsymbol{q}(t)-\boldsymbol{r}_{k}^{j}), & \forall t \in\left[0, T_{k}\right].
    \end{aligned}\right.
\end{equation}
Inspired by \cite{ref12}, we transform \(\mathcal{G}_{\star}\) into finite
inequality constraints using the integral of constraint violations. which is furthered transformed into the penalized sampled function \(\mathcal{J}_{\star}\).
\begin{subequations}\label{eq13}
\begin{align}
\mathcal{I}_{k}^{\star} & =\frac{T_{k}}{L} \sum_{l=0}^{L} \bar{\omega}_{l} \max \left[\mathcal{G}_{\star}\left(\boldsymbol{c}_{k}, T_{k}, \frac{l}{L}\right), \mathbf{0}\right]^{3}, \label{eq13A} \\
\mathcal{J}_{\star} & =\sum_{k=1}^{K} \mathcal{I}_{k}^{\star}, \star=\{v,\omega,c\}, \label{eq13B}\\
\frac{\partial \mathcal{J}_{\star}}{\partial c_{k}} & =\frac{\partial \mathcal{I}_{k}^{\star}}{\partial \mathcal{G}_{\star}} \frac{\partial \mathcal{G}_{\star}}{\partial c_{k}}, \frac{\partial \mathcal{J}_{\star}}{\partial T_{k}}=\frac{\mathcal{I}_{k}^{\star}}{T_{k}}+\frac{l}{L} \frac{\partial \mathcal{I}_{k}^{\star}}{\partial \mathcal{G}_{\star}} \frac{\partial \mathcal{G}_{\star}}{\partial t}, \label{eq13C}\\
\frac{\partial \mathcal{I}_{k}^{\star}}{\partial \mathcal{G}_{\star}} & =3 \frac{T_{k}}{L} \sum_{l=0}^{L} \bar{\omega}_{l} \chi^{\mathrm{T}} \max \left[\mathcal{G}_{\star}\left(\mathbf{c}_{k}, T_{k}, \frac{l}{L}\right), \mathbf{0}\right]^{2},\label{eq13D}
\end{align}
\end{subequations}
where \(L\) controls the numerical integration accuracy, \(\chi^{\mathrm{T}}\) represents the penalty weight vector, \(\left(\bar{\omega}_{0},\bar{\omega}_{1},...,\bar{\omega}_{L-1},\bar{\omega}_{L}\right)=\left(1/2,1,...,1,1/2\right)\) are the quadrature coefficients following the trapezoidal rule \cite{ref20}.

\section{Motion Control For Morphing Quadrotors}
\subsection{Nonlinear Controller}
\begin{figure}[!t]
    \hspace{-0.1cm}\includegraphics[width=0.5\textwidth]{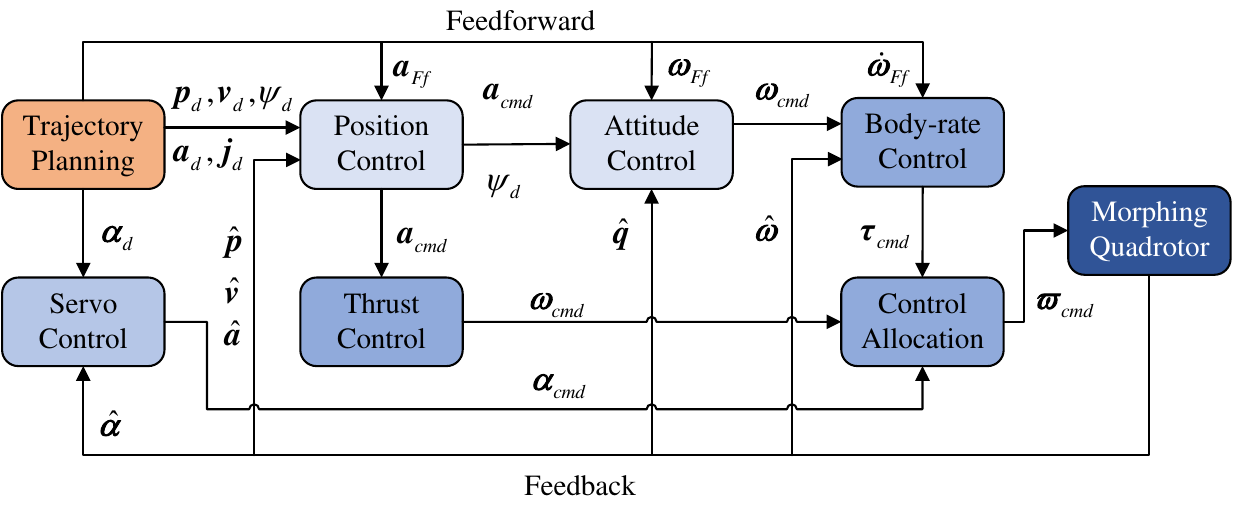}
    \caption{Nonlinear cascaded control framework for a morphing quadrotor.}
    \label{fig5}
    \vspace{-0.3cm}
\end{figure}

As discussed in Sect. II, previous methods \cite{ref3,ref4} didn't consider the influence of nonlinear aerodynamic drag and relied on linearized dynamic models in real-world experiments. 
To ensure stable and precise trajectory tracking of the morphing quadrotor during morphing. 
The design of the nonlinear controller mainly consists of three components: feedforward control, feedback control, and online estimation of thrust coefficient.
The adaptive cascaded control framework for the morphing quadrotor is shown in Fig. \ref{fig5}.
\subsubsection{\textbf{Control Law Design}}
Firstly, the reference input \(\left\{\boldsymbol{p}_{d},\boldsymbol{v}_{d},\boldsymbol{a}_{d},\boldsymbol{j}_{d},{{\psi}_{d}}\right\}\) of the controller is determined through trajectory planning. For the position controller, we need to calculate the desired acceleration \(\boldsymbol{a}_{cmd}\) based on the desired position, velocity, acceleration, and feedforward acceleration \(\boldsymbol{a}_{Ff}\) influenced by aerodynamic drag, as shown below:
\begin{equation}
    \label{eq14}
    \boldsymbol{a}_{cmd}=\boldsymbol{K}_{a}\boldsymbol{a}_{d}+\boldsymbol{a}_{Ff}-\boldsymbol{a}_{Fb}+\boldsymbol{g}\boldsymbol{z}_{E},
\end{equation}
where \(\boldsymbol{K}_{a}\) is a constant matrix, \(\boldsymbol{a}_{Fb}\) represents the sum of the proportional term of position error and the integral term of velocity error and \(\boldsymbol{g}= \left[0,0,-g \right]^\mathrm{T}\) is the gravitational vector. Furthermore, we utilize the obtained desired acceleration \(\boldsymbol{a}_{cmd}\) to estimate the thrust \({f}_{cmd}\), and use the \(\boldsymbol{a}_{cmd}\) and desired yaw angle \({\psi}_{d}\) as inputs to the attitude controller.

We represent the attitude angles using the rotation matrix \(\boldsymbol{R}_{d}\), and constrain the projection of \(\boldsymbol{x}^{B}\) onto \(\boldsymbol{x}^{E}-\boldsymbol{y}^{E}\) to coincide with \(\boldsymbol{x}^{C}\). Consequently, \(\boldsymbol{y}^{C}\) can be expressed as follows:
\begin{equation}
    \label{eq15}
    \boldsymbol{y} ^{C} = \left [ -\sin {\psi_{d}}, \cos {\psi_{d}, 0} \right ]^\mathrm{T},  
\end{equation}
then the body rotation matrix \(\boldsymbol{R}_{d}=\left[ \boldsymbol{x}_{d}^{B},\boldsymbol{y}_{d}^{B},\boldsymbol{z}_{d}^{B}\right]\) can be represented as:
\begin{equation}
    \label{eq16}
    \boldsymbol{R}_{d}=\left[\frac{\boldsymbol{y}^{C} \times \boldsymbol{z}_{d}^{B}}{\left\|\mathbf{y}^{C} \times \boldsymbol{z}_{d}^{B}\right\|},    \boldsymbol{z}_{d}^{B} \times \boldsymbol{x}_{d}^{B},  \frac{\boldsymbol{a}_{c m d}}{\left\|\boldsymbol{a}_{c m d}\right\|}\right],
\end{equation}
we represent the rotation matrix \(\boldsymbol{R}_{d}\) using quaternion \(\boldsymbol{q}_{d}\). Then,
we calculate the desired body angular velocity \(\boldsymbol{\omega}_{cmd}\) as follows:
\begin{equation}
    \label{eq17}
    \boldsymbol{\omega}_{cmd}=\boldsymbol{K}_{A} \operatorname{sgn}\left(q_{e, 0}\right) \boldsymbol{q}_{e, 1: 3}+\boldsymbol{\omega}_{Ff}, \quad \boldsymbol{q}_{e}=\hat{\boldsymbol{q}}^{-1} \boldsymbol{q}_{d},
\end{equation}
where \(\boldsymbol{K}_{A}\) is a constant matrix, \(\boldsymbol{q}_{e}\) represents the rotation from the state feedback quaternion \(\hat{\boldsymbol{q}}\) to the desired attitude \(\boldsymbol{q}_{d}\), \(\operatorname{sgn}\left(.\right)\) denotes the sign function, and \(\boldsymbol{\omega}_{Ff}\) represents the feedforward body angular velocity obtained through differential flatness. Then, according to equation \eqref{eq1E}, we can obtain the expressions for the three-axis torque as follows:
\begin{equation}
    \label{eq18}
    \boldsymbol{\tau}=\boldsymbol{J}\left(\dot{\boldsymbol{\omega}}_{Ff}+\dot{\boldsymbol{\omega}}_{cmd}\right)+\boldsymbol{\omega}_{cmd} \times \boldsymbol{J} \boldsymbol{\omega}_{cmd}+\boldsymbol{A} \boldsymbol{R}^\mathrm{T}_{d} \boldsymbol{v}+\boldsymbol{B} \boldsymbol{\omega}_{cmd}.
\end{equation}
Where \(\dot{\boldsymbol{\omega}}_{Ff}\) represents the reference angular acceleration obtained through feedforward computation, \(\dot{\boldsymbol{\omega}}_{cmd}\) is obtained through Proportional-Integral-Derivative (PID) operations to process the deviation between the desired body angular velocity and the feedback body angular velocity.

\subsubsection{\textbf{Thrust Coefficient Online Estimation}}
Due to the overlap between the propellers or between the center body and the propellers after folding, there may be some power loss for the morphing quadrotor. To address this issue, we improve the structure of the drone to reduce the extent of overlap in various forms. Furthermore, we design a thrust estimation scheme based on the forgetting factor recursive least squares method \cite{ref21}, as detailed below:
\begin{equation}
    \label{eq19}
    \left\{\begin{aligned}
    K_{n}&=P_{n-1} c_{n}\left(\rho+c_{n}^\mathrm{T} P_{n-1} c_{n}\right)^{-1}, \\
    P_{n}&=\left(1-K_{n} c_{n}^\mathrm{T}\right) P_{n-1} / \rho, \\
    H_{n}&=H_{n-1}+K_{n}\left(\hat{a}_{z}-c_{n} H_{n-1}^\mathrm{T}\right), \\
    c_{n}&=a_{c m d,z} / H_{n},
    \end{aligned}\right.
\end{equation}
where \(K\) represents the model gain, \(P\) is a measurable constant, \(\rho\) is a constant related to IMU vibrations during flight, and \(c\) is the identified normalized thrust. \(H\) denotes the slope of the real-time estimated thrust model concerning the desired \(z\)-axis acceleration.
\vspace{-0.5cm}
\subsection{Servo Controller}

Deformation control of the morphing quadrotor is primarily achieved by rotating servos to specified positions. The calculated desired angle \(\boldsymbol{\alpha}_{d}\) from trajectory planning is input into the servo controller. The angle deviation is denoted as \(\boldsymbol{e}_{\alpha}=\boldsymbol{\alpha}_{d}-\hat{\boldsymbol{\alpha}}\), and the following control law can be designed:
\begin{equation}
    \label{eq20}
    \boldsymbol{\alpha}_{cmd}=\boldsymbol{K}_{\alpha p}\boldsymbol{e}_{\alpha} + \boldsymbol{K}_{\alpha d} \dot{\boldsymbol{e}}_{\alpha},
\end{equation}
where \(\boldsymbol{K}_{\alpha p}\) and \(\boldsymbol{K}_{\alpha d}\) represent the proportional and differential constants of the servo controller.

\begin{figure}[!t]
    \hspace{0.5cm}\includegraphics[width=0.4\textwidth]{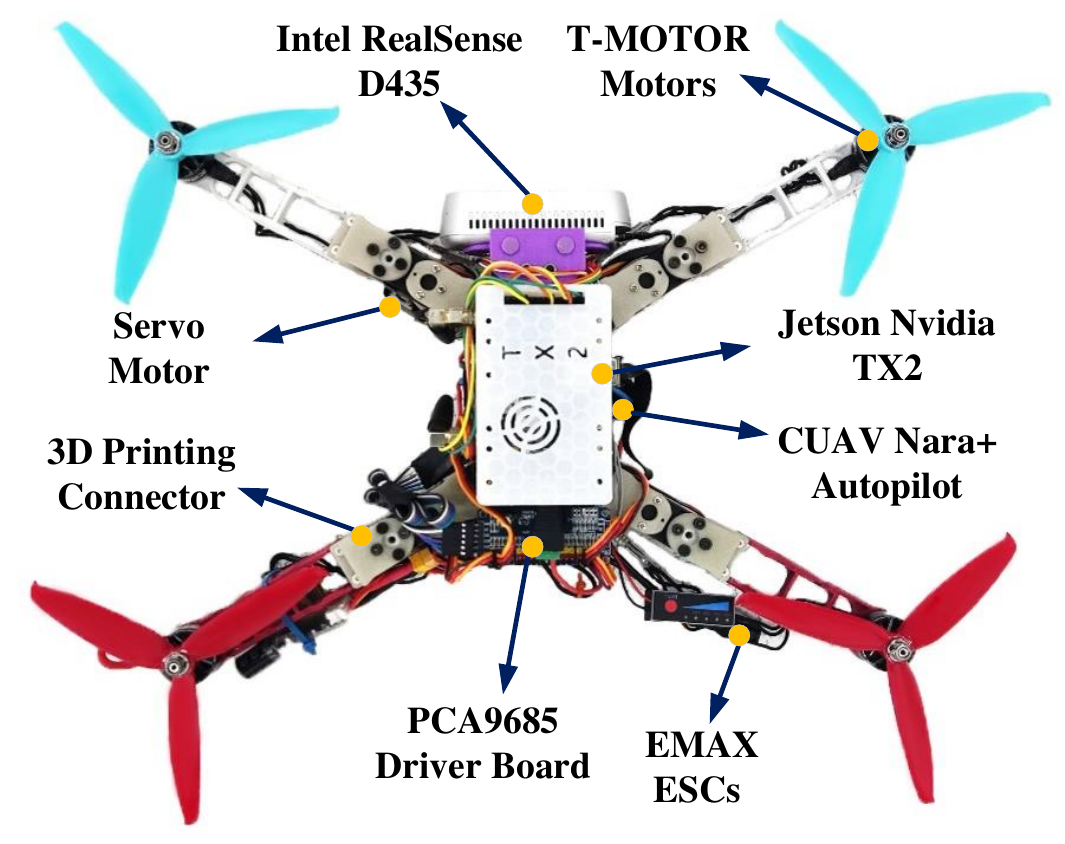}
    \caption{The detailed composition of the morphing quadrotor platform.}
    \label{fig6}
    \vspace{-0.1cm}
\end{figure}

\begin{table}[!h]
    \renewcommand{\arraystretch}{1.3}
    \caption{Description of The Main Components}
    \label{tab:table1}
    \centering
    \begin{tabular}{|c|c|l|}
        \hline
        Components & Part numbers & Key features \\
        \hline
        Onboard computer & Jetson Nvidia TX2 & GPU: 256 CUDA cores \\
        \hline
        Flight controller & CUAV Nara+ & CPU: STM32H743 \\
        \hline
        Camera & Intel RealSense D435 & FOV: \(\SI{85.2}{\degree} \times \SI{58}{\degree}\)\\
        \hline
        Motor & T-Motor F60 & KV: \SI{2550}{\kilo\volt}  \\
        \hline
        ESCs & EMAX 45A & MCU: STM32F051 \\
        \hline
        Propeller & Flash 6042 & Diameter: 6 inch \\
        \hline
        Servo & RDS3115 & Dead band width: \SI{3}{\micro\second} \\
        \hline
        Driver board & PCA9685 & Resolution: 12-bit \\
        \hline
        Battery & GS33004S30 & Weight: \SI{235}{\gram} \\
        \hline
        Robot  &  Customized aircraft & Size: \(48\times 40\times \)\SI{10}{\centi\meter}\\ 
        \hline
    \end{tabular}
\end{table}

\section{Experiments Results}
\vspace{0.3cm}
\subsection{Morphing Quadrotor Platform}

Our morphing quadrotor, as shown in Fig. \ref{fig6}, mainly consists of three modules.
The flight module comprises a flight controller, four ESCs, and four motors that generate thrust by rotating the propellers. The transformation module includes a servo driver board, four 3D-printed connectors, and four servos for arm rotation. The motion planning and control module integrates an onboard computer and camera, with the computer handling control, planning, localization, and mapping. The description of the main components is listed in Tab. \ref{tab:table1}.




\subsection{Software Architecture}

The architecture of our proposed planning and control framework is shown in Fig. \ref{fig2}. Firstly, the open-source VINS-Fusion\cite{ref22} framework is adopted for visual localization, using stereo images (\SI{15}{\hertz}) and IMU data (\SI{200}{\hertz}) as inputs to offer Visual  Inertial Odometry (VIO). Secondly, a global point cloud map is pre-constructed on the onboard computer to ensure high-quality map construction in narrow spaces. In the planning stage, the weighted A* algorithm \cite{ref16} is used to search for collision-free paths and construct the SFC. Subsequently, the desired trajectory is obtained by spatial-temporal trajectory optimization. Finally, the designed nonlinear controller is employed to track the desired trajectory.

\begin{table}[!t]
\caption{Trajectory Tracking Error of PID, LQR and Proposed controller\label{tab:table2}}
\centering
\setlength\tabcolsep{3pt}
\renewcommand\arraystretch{1.5}
\begin{tabular}{|c|c|c|c|c|c|c|}
\hline
\multicolumn{1}{|c|}{\multirow{2}{*}{\({v_{max}}\)[m/s]}} & \multicolumn{3}{c}{Error (average) {[}m{]}} & \multicolumn{3}{|c|}{Error (max) {[}m{]}} \\
\cline{2-7}
\multicolumn{1}{|c|}{} & PID\cite{ref4} & LQR\cite{ref3} & Proposed  & PID\cite{ref4}  & LQR\cite{ref3} & Proposed    \\
\hline
0.6  & 0.0248  & 0.0179  & \textbf{0.0097}   & 0.0986   & 0.0765  & \textbf{0.0325}  \\
\hline
0.8  & 0.0364  & 0.0253  & \textbf{0.0164}   & 0.1247   & 0.0964  & \textbf{0.0596}  \\
\hline
1.0  & 0.0482  & 0.0305  & \textbf{0.0241}   & 0.1541   & 0.1265  & \textbf{0.0987} \\
\hline
\end{tabular}
\end{table}

\begin{figure}[!t]
    \hspace{-0.35cm}\includegraphics[width=0.5\textwidth]{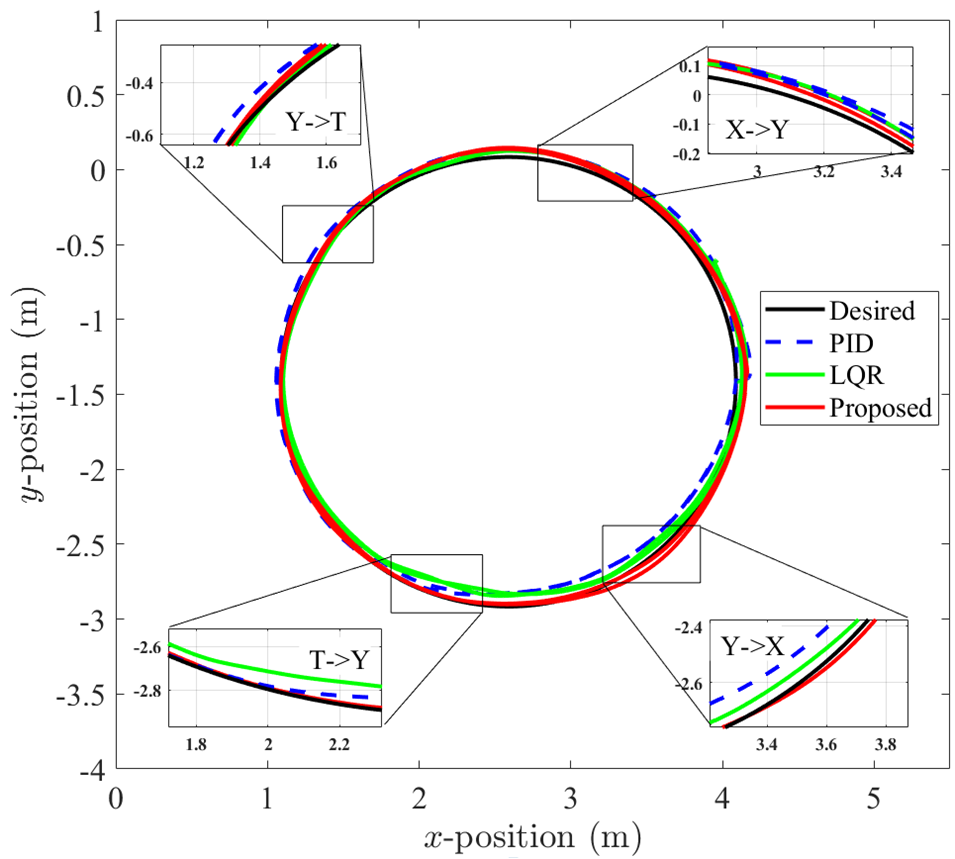}
    \caption{Benchmark of tracking circle trajectories at a maximum velocity of \SI{1.0}{m/s} while constantly morphing. More details are included in the video.}
    \label{fig7}
    \vspace{-0.4cm}
\end{figure}

\subsection{Benchmark Comparison}

We compare the proposed controller with PID \cite{ref4} and LQR \cite{ref3} controller in real-world environments. The morphing quadrotor uses each controller to track a circular trajectory with different velocities while constantly morphing. As shown in Tab. \ref{tab:table2} and Fig. \ref{fig7}, our proposed method has minor tracking errors compared to other methods. Further analysis is as follows. Derrouaou \textit{et al}. \cite{ref4} can't track the angular velocity well due to changes in the inertia matrix during deformation, which affects the tracking performance of the outer loop position. On the other hand, Falanga \textit{et al}. \cite{ref3} and our proposed method can both adaptively adjust the inertia, but the former design is based on model linearization and doesn't take into account the impact of nonlinear aerodynamic drag on quadrotor tracking performance. It can't handle motor saturation, especially at high speeds.

In summary, the proposed controller is more suitable for trajectory tracking of morphing quadrotors among the above controllers. Firstly, our controller for attitude calculation doesn't rely on the small-angle assumption. Then, we account for nonlinear aerodynamic drag to avoid significant distortion of the linearized model. Detailed tracking experiments are presented in Sect. VI-E.

\subsection{Simulation Experiments}
\begin{figure*}[!t]
    \hspace{0.0cm}\includegraphics[width=1\textwidth]{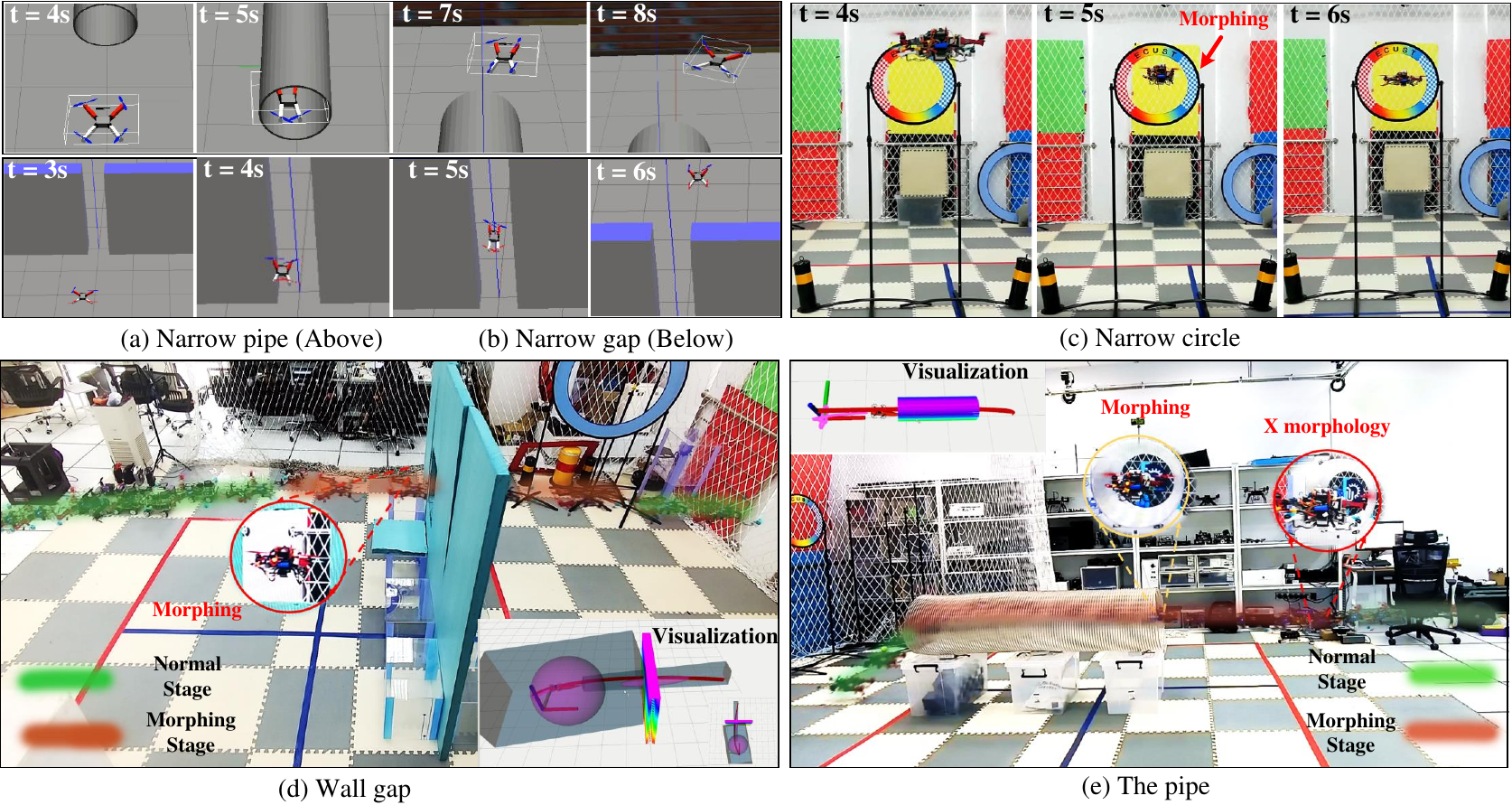}
    \vspace{-0.7cm}
    \caption{Flight results in Gazebo simulations and real-world experiments. (a) shows the process of traveling the pipe horizontally. (a) shows the process of crossing the gap horizontally in the simulation environment. (c) shows the process of crossing a narrow circle by deforming at different time instances. (d) shows the process of traveling the wall gap, which consists of two stages, namely the normal stage (green) and the morphing stage (orange). (e) shows the process of crossing the pipe in the real world. More details are included in the video.}
    \label{fig8}
    \vspace{-0.3cm}
\end{figure*}



In the Gazebo simulation environment, we construct a simplified version of the morphing quadrotor SDF model and perform simulation experiments for traversing the pipe and narrow gap, as shown in Fig. \ref{fig8}(a-b). 

\subsubsection{\textbf{Traversing Narrow Pipe}}
We set the inner diameter of the pipe entrance to \SI{50}{\centi\meter}, with a length of \SI{400}{\centi\meter}. As shown in Fig. \ref{fig8}(a), our morphing quadrotor can successfully traverse the pipe, even under long-distance deformation conditions.

\subsubsection{\textbf{Crossing Narrow Gap}}
We set the horizontal distance of the gap to \SI{50}{\centi\meter}, while the horizontal width of the quadrotor in X configuration is \SI{60}{\centi\meter}. After morphing, the width of the quadrotor is reduced to \SI{40}{\centi\meter} (including the propellers). As shown in Fig. \ref{fig8}(b), the experiment shows that the morphing quadrotor can safely traverse the narrow gap.

\subsection{Real Experiments}


One of the advantages of morphing quadrotors over conventional quadrotors is that can reduce the horizontal width by folding their arms (up to a maximum reduction of {\bf{37.5\(\%\))}} to adapt to different environments. To validate the scene-adaptive ability of the morphing quadrotor in various restricted environments and evaluate the effectiveness of the proposed framework, we construct a diverse set of restricted scenarios including narrow circle, wall gap, pipe, and complex restricted environments in real-world experiments. More details are included in the video.
\subsubsection{\textbf{Crossing Narrow Circle}}
The inner diameter of the circle is \SI{40}{\centi\meter}, while the horizontal width of the quadrotor is \SI{48}{\centi\meter}. The maximum speed is set to \SI{1.25}{m/s} Even the maximum safe distance doesn't exceed \SI{5}{\centi\meter}, the quadrotor is still capable of stable deformation and successful traversal through the narrow circle. Fig. \ref{fig8}(c) shows the moments as the quadrotor passes through the circle.

\subsubsection{\textbf{Traversing Wall Gap}}
We set a square opening with a side length of \SI{40}{\centi\meter}. In this experiment, we consider two settings: one similar to a conventional quadrotor flying over the high wall to complete the task, and the other traversing through the gap by deformation, as shown in Fig. \ref{fig8}(d). We limit the maximum speed for deformation traversal to \SI{2.5}{m/s}. In the visualization section, our quadrotor can safely fly along the trajectory and collision-free in SFC. Based on the battery consumption during the flight, we calculate the percentage of battery power used during the two flights. The result shows that deformation traversal through the wall gaps can save {\bf{one-third}} of the energy compared to flying over the high wall. This is mainly due to the latter requiring greater acceleration to accomplish the task.

\subsubsection{\textbf{Crossing Narrow Pipe}}
We take on the challenge of long-distance traverse through a more difficult pipe. The pipe has a length of \SI{190}{\centi\meter} and an inner diameter width of \SI{45}{\centi\meter}, with a maximum speed of \SI{3.5}{m/s}. Compared to \cite{ref23}, which conducted manual experimental tests for ground-based deformation traversal by the terrestrial-aerial quadrotor. Our quadrotor relies on onboard localization and computation to traverse the pipe at varying heights, as shown in Fig. \ref{fig8}(e). This experiment effectively validates the robustness of our proposed framework.
\begin{figure}[!t]
    \hspace{0.05cm}\includegraphics[width=0.48\textwidth]{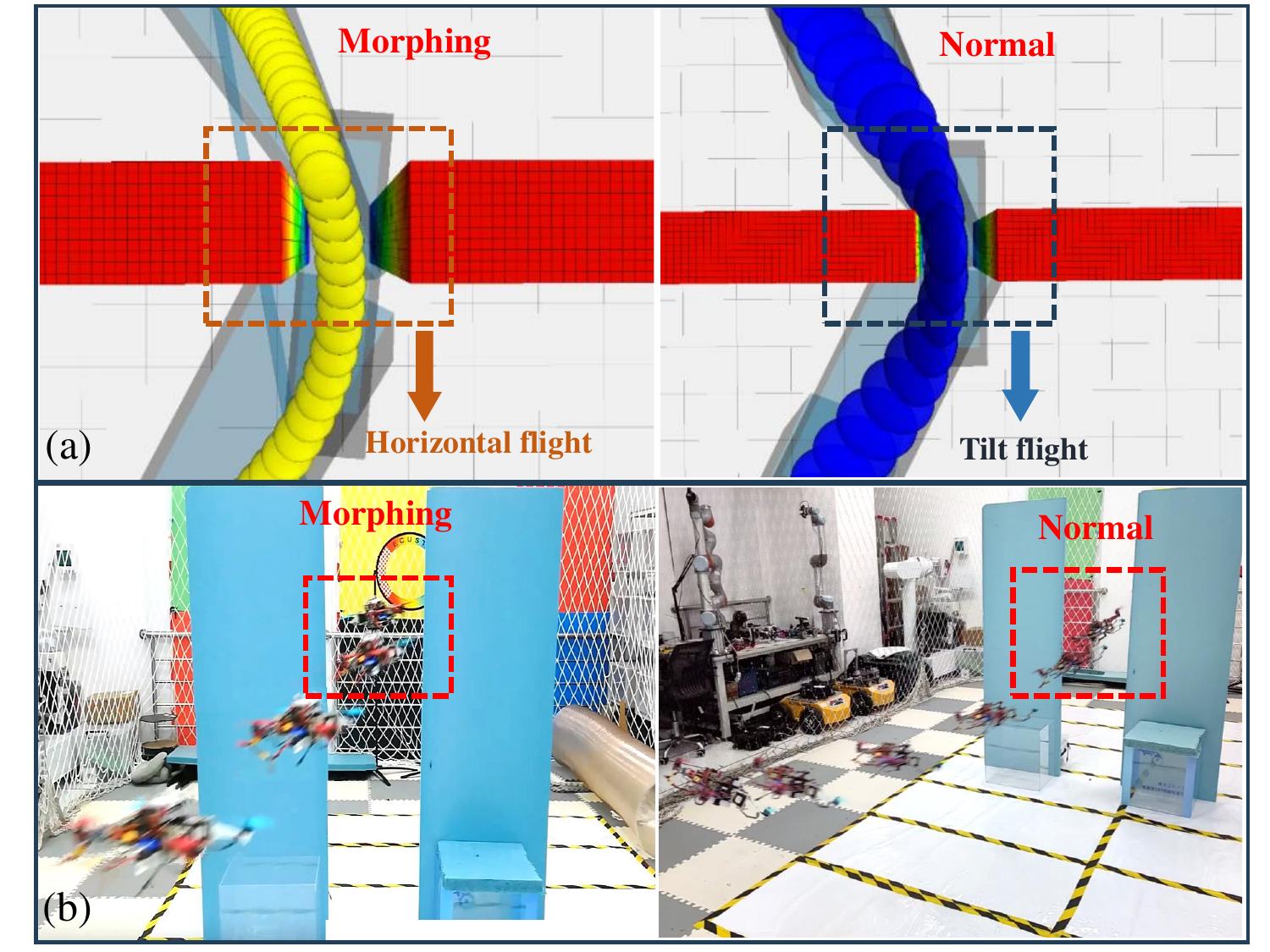}
    \caption{Comparison of trajectories through a narrow gap in simulation and real-world experiments. The ellipsoids depict the position and attitude of the quadrotor. Slightly inclined horizontal trajectories are produced by accounting for deformation on the left side of the figure, while highly inclined trajectories are generated without considering deformation on the right side of the figure.}
    \label{fig9} 
    \vspace{-0.4cm}
\end{figure}

\subsubsection{\textbf{Traversing Complex and Restricted Environment}}
The complex restricted environment is composed of two circles with inner diameters of \SI{45}{\centi\meter} and \SI{80}{\centi\meter}, as well as a square notch with a side length of \SI{40}{\centi\meter}. We set the maximum speed to \SI{2.5}{m/s}. The quadrotor must undergo adaptive deformation to traverse these restricted objects, which brings significant challenges to planning and control. 
The flight trajectory is shown in Fig. \ref{fig1}(a), where our quadrotor can smoothly and quickly navigate through narrow gaps. 
A similar work is \cite{ref12} which proposed a planning method based on SE(3) and performed traversing multiple narrow gaps. However, it requires a significant amount of space for acceleration and deceleration (similar to the right side in Fig. \ref{fig9}), making it difficult to implement in confined spaces. In contrast, Our morphing quadrotor is capable of maneuvering horizontally narrow gaps with minimal acceleration changes, as shown on the left side in Fig. \ref{fig9}. Our planning method provides a novel solution for traversing complex restricted environments.
\section{Conclusion}
In this paper, we propose a framework for morphing quadrotors to perform autonomous flight in complex restricted environments. The framework mainly consists of a full-body trajectory optimizer based on SFC and a unified nonlinear controller that accounts for aerodynamic drag. In addition, we delegate the computational tasks of the planning, control, mapping, and localization modules entirely to the onboard platform of the quadrotor. We have conducted numerous experiments in challenging simulations and real-world environments to demonstrate the performance and robustness of the proposed framework.

For future work, our primary goal is to achieve autonomous navigation through morphing in unknown and restricted environments. Furthermore, we will fully exploit the adaptive morphological modifications advantage of the quadrotor, allowing it to autonomously perform tasks such as object grasping and close-range detection on vertical surfaces.

\newpage

\vfill


\begin{thebibliography}{1}
\bibliographystyle{IEEEtran}







\bibitem{ref1}
J. Delmerico, S. Mintchev, A. Giusti, B. Gromov, K. Melo, T. Horvat, C . Cadena, M. Hutter, A. Ijspeert, D. Floreano, et al., ``The current state and future outlook of rescue robotics,'' \textit{Journal of Field Robotics}, vol . 36, no. 7, pp. 1171–1191, 2019.
\bibitem{ref2}
Q. Sun, J. Fang, W. X. Zheng and Y. Tang, ``Aggressive Quadrotor Flight Using Curiosity-Driven Reinforcement Learning,'' in \textit{IEEE Transactions on Industrial Electronics}, vol. 69, no. 12, pp. 13838-13848, Dec. 2022.
\bibitem{ref3}
D. Falanga, K. Kleber, S. Mintchev, D. Floreano, and D. Scaramuzza, ``The foldable drone: A morphing quadrotor that can squeeze and fly,'' \textit{ IEEE Robotics and Automation Letters}, vol. 4, no. 2, pp. 209–216, 2019.

\bibitem{ref4}
Derrouaoui, S.H., Bouzid, Y. \& Guiatni, M., ``PSO Based Optimal Gain Scheduling Backstepping Flight Controller Design for a Transformable Quadrotor.'' \textit{J Intell Robot Syst}, 102, 67 (2021).

\bibitem{ref5}
A. Papadimitriou, S. S. Mansouri, C. Kanellakis and G. Nikolakopoulos, ``Geometry Aware NMPC Scheme for Morphing Quadrotor Navigation in Restricted Entrances,'' \textit{ 2021 European Control Conference}, Delft, Netherlands, 2021, pp. 1597-1603.

\bibitem{ref6}
J. M. Butt, X. Ma, X. Chu and K. W. Samuel Au, ``Adaptive Flight Stabilization Framework for a Planar 4R-Foldable Quadrotor: Utilizing Morphing to Navigate in Confined Environments,'' \textit{2022 American Control Conference}, Atlanta, GA, USA, 2022, pp. 1-7.


\bibitem{ref7}
N. Bucki, J. Tang, M. W. Mueller, ``Design and Control of a Midair-Reconfigurable Quadcopter using Unactuated Hinges,'' in \textit{IEEE Transactions on Robotics}, pp. 539-557, 2023.

\bibitem{ref8}
V. Riviere, A. Manecy, and S. Viollet, ``Agile robotic fliers: A morphing-based approach,'' \textit{Soft robotics}, vol. 5, no. 5, pp. 541–553, 2018.

\bibitem{ref9}
Kose, O. and Oktay, T., ``Simultaneous quadrotor autopilot system and collective morphing system design, '' \textit{Aircraft Engineering and Aerospace Technology},  Vol. 92 No. 7, pp. 1093-1100, 2020.

\bibitem{ref10}
C. Kim, H. Lee, M. Jeong and H. Myung, ``A Morphing Quadrotor that Can Optimize Morphology for Transportation,'' \textit{ 2021 IEEE/RSJ International Conference on Intelligent Robots and Systems}, Prague, Czech Republic, 2021, pp. 9683-9689.

\bibitem{ref11}
Y. Wu et al., ``Ring-Rotor: A Novel Retractable Ring-Shaped Quadrotor With Aerial Grasping and Transportation Capability,'' in \textit{IEEE Robotics and Automation Letters}, vol. 8, no. 4, pp. 2126-2133, 2023.

\bibitem{ref12}
Z. Wang, X. Zhou, C. Xu, and F. Gao, ``Geometrically Constrained Trajectory Optimization for Multicopters''. \textit{IEEE Transactions on Robotics}, vol. 38, no. 5, pp. 3259-3278, 2022.

\bibitem{ref13}
A. Fabris, K. Kleber, D. Falanga and D. Scaramuzza, ``Geometry-aware Compensation Scheme for Morphing Drones,'' \textit{2021 IEEE International Conference on Robotics and Automation},  Xi'an, China, 2021, pp. 592-598.


\bibitem{ref14}
D. Hu, Z. Pei, J. Shi, and Z. Tang, ``Design, modeling and control of a novel morphing quadrotor,'' \textit{IEEE Robotics and Automation Letters}, vol. 6, no. 4, pp. 8013-8020, 2021.

\bibitem{ref15}
M. Faessler, A. Franchi and D. Scaramuzza, ``Differential Flatness of Quadrotor Dynamics Subject to Rotor Drag for Accurate Tracking of High-Speed Trajectories,'' in \textit{IEEE Robotics and Automation Letters},  vol. 3, no. 2, pp. 620-626, 2018.



\bibitem{ref16}
I. Pohl, ``Heuristic search viewed as path finding in a graph,'' \textit{Artificial Intelligence}, vol. 1, no. 3–4, pp. 193–204, 1970.



\bibitem{ref17}
F. Gao, W. Wu, W. Gao, and S. Shen, ``Flying on point clouds: Online trajectory generation and autonomous navigation for quadrotors in cluttered environments.'' \textit{Journal of Field Robotics}, 36(4):710–733, 2019.

\bibitem{ref18}
J. Ji, Z. Wang, Y. Wang, C. Xu, and F. Gao. ``Mapless-planner: A robust and fast planning framework for aggressive autonomous flight without map fusion.'' in \textit{2021 IEEE International Conference on Robotics and Automation}, 2021, pp. 6315–6321.

\bibitem{ref19}
S. Liu, M. Watterson, K. Mohta, K. Sun, S. Bhattacharya, C. J. Taylor, and V. Kumar, ``Planning dynamically feasible trajectories for quadrotors using safe flight corridors in 3-d complex environments.'' in \textit{IEEE Robotics and Automation Letters}, 2(3):1688–1695, 2017.


\bibitem{ref20}
W. H. Press, S. A. Teukolsky, W. T. Vetterling, and B. P. Flannery, \textit{Numerical Recipes with Source Code CD-ROM 3rd Edition: The Art
of Scientific Computing.} Cambridge University Press, 2007.

\bibitem{ref21}
C. Paleologu, J. Benesty, and S. Ciochina, ``A robust variable forgetting factor recursive least-squares algorithm for system identification,'' \textit{IEEE Signal Process. Lett.}, vol. 15, no. 9, pp. 597–600, 2008.

\bibitem{ref22}
T. Qin, P. Li, and S. Shen, ``Vins-mono: A robust and versatile monocular visual-inertial state estimator,''. \textit{IEEE Transactions on Robotics}, vol. 34, no. 4, pp. 1004–1020, 2018.

\bibitem{ref23}
E. Gefen and D. Zarrouk, ``Flying STAR2, a Hybrid Flying Driving Robot With a Clutch Mechanism and Energy Optimization Algorithm,'' in \textit{IEEE Access}, vol. 10, pp. 115491-115502, 2022.

\end{thebibliography}
\end{document}